\documentclass[conference]{IEEEtran}
\IEEEoverridecommandlockouts
\usepackage{booktabs}
\usepackage[table]{xcolor}
\usepackage{graphicx}
\usepackage{subfigure}
\usepackage{stfloats}
\usepackage{textcomp}
\usepackage{balance}
\usepackage{amsmath}
\usepackage{amsfonts,amssymb}
\usepackage{bm}
\usepackage{extarrows}
\usepackage[colorlinks=true,
            linkcolor=blue,
            anchorcolor=black,
            citecolor=black,
            ]{hyperref}
\usepackage[margin=1.9cm]{geometry}
\usepackage[linesnumbered,ruled,vlined]{algorithm2e}
\usepackage{cite}
\usepackage{url}
\usepackage{multirow}  
\usepackage{array}    
\title{\LARGE \bf
IGV-RRT: Prior-Real-Time Observation Fusion for Active Object Search in Changing Environments
}

\author{Wei Zhang$^{1,2,\dag}$, Ping Gong$^{1,\dag}$, Yujie Wang$^{1}$, Leilei Yao$^{1}$, Minghui Bai$^{1}$, Rongfeng Ye$^{1}$, Yinchuan Wang$^{1}$, \\Yachao Wang$^{1}$, Chen Sun$^{2}$, Chaoqun Wang$^{1,*}$

\thanks{$^{1}$The School of Control Science and Engineering, Shandong University}
\thanks{$^{2}$Department of Data and Systems Engineering, HKU}
\thanks{$^{*}$Corresponding author. 
Email: \tt chaoqunwang@sdu.edu.cn}%

\thanks{$^{\dag}$The first two authors contributed equally to this work.}%
}

\begin{document}
\nocite{*}
\newgeometry{left=1.9cm, right=1.9cm, top=2.54cm, bottom=1.9cm}
\maketitle
\thispagestyle{empty}
\pagestyle{empty}
\begin{abstract}
Object Goal Navigation (ObjectNav) in temporally changing indoor environments is challenging because object relocation can invalidate historical scene knowledge. To address this issue, we propose a probabilistic planning framework that combines uncertainty-aware scene priors with online target relevance estimates derived from a Vision Language Model (VLM). The framework contains a dual-layer semantic mapping module and a real-time planner. The mapping module includes an Information Gain Map (IGM) built from a 3D scene graph (3DSG) during prior exploration to model object co-occurrence relations and provide global guidance on likely target regions. It also maintains a VLM score map (VLM-SM) that fuses confidence-weighted semantic observations into the map for local validation of the current scene. Based on these two cues, we develop a planner that jointly exploits information gain and semantic evidence for online decision making. The planner biases tree expansion toward semantically salient regions with high prior likelihood and strong online relevance through IGV-RRT, while preserving kinematic feasibility during online planning. Simulation and real-world experiments demonstrate that the proposed method effectively mitigates the impact of object rearrangement, achieving higher search efficiency and success rates than representative baselines in complex indoor environments.
\end{abstract}


\section{Introduction}
Reliable target search in temporally changing indoor environments remains a fundamental challenge, especially for robots that operate over long periods in previously visited spaces. To provide prompt service, the robot operating in an indoor environment accumulates knowledge about the environment, including room layouts, objects, and their context relations. Such historical experience can provide valuable global guidance for the search. However, indoor environments are rarely static over time: objects may be moved, occluded, or rearranged, causing a mismatch between historical knowledge and the current scene. Therefore, for ObjectNav \cite{sun2024survey} in long-term deployments, a central challenge is how to effectively exploit accumulated historical experience while remaining adaptive to temporal scene evolution, so that target search can be performed efficiently and reliably.

\begin{figure}[t]
    \centering
    \includegraphics[width=1\linewidth]{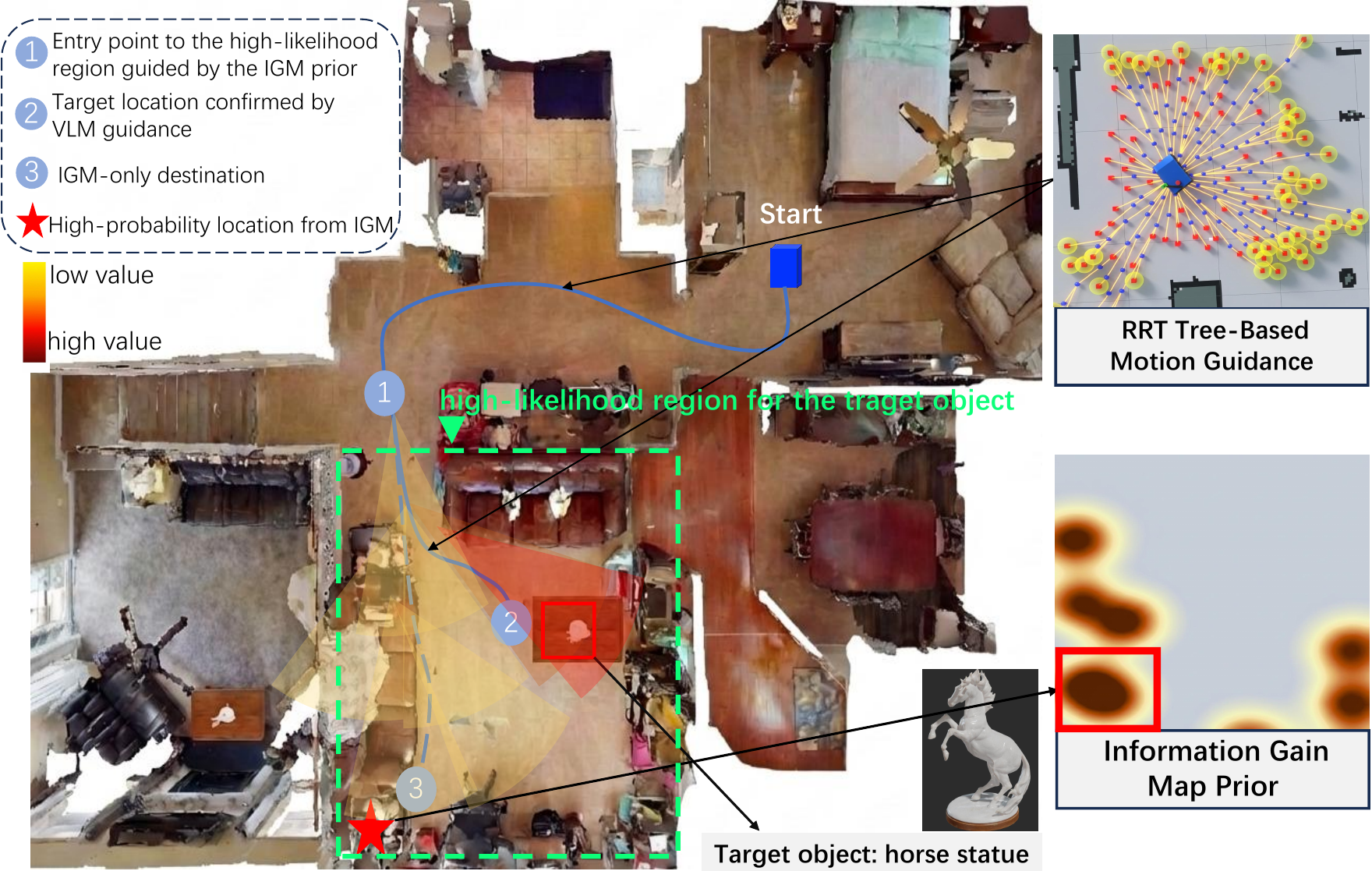}
    \caption{Active object search in a time-varying indoor scene. The static IGM prior guides global navigation toward a high-likelihood region for the target. Online observations are processed by BLIP-2 and fused into a VLM-SM to refine local motion toward the true target. The IGM-only endpoint indicates prior bias.}
    \label{fig: compare}
\end{figure}

Long-horizon object goal navigation in temporally changing indoor environments inherently relies on the coordination of multiple information sources. Historical environmental knowledge accumulated from previous exploration can provide coarse global guidance for target search. Commonsense knowledge embedded in foundation models further supports inferring plausible target regions from semantic context. After entering a plausible target region, the robot leverages real-time local observations to acquire up-to-date evidence of the current object distribution and scene state, thereby mitigating the impact of stale historical knowledge caused by temporal changes and ensuring robust and efficient target object search.

Existing methods exploit these sources in different ways. One class of approaches introduces historical knowledge in advance to bias navigation, for example, through offline-constructed probability map \cite{yang2018visual,zhang2019efficient} or by leveraging 3DSG to infer likely target locations \cite{gassol2026relationship,zhou2025fsr}. Such methods can incorporate environment structure and contextual relations into search and reduce invalid exploration. Nonetheless, such priors are often constructed in an offline manner and remain fixed during deployment. As indoor environments change over time, objects can be relocated or reconfigured, which may cause previously reliable historical knowledge to become stale and potentially misdirect the navigation policy. This issue is particularly pronounced for methods that rely on explicit 3D scene representations, as constructing and maintaining an accurate scene graph under object displacement is a challenging task in itself.
\begin{figure*}[t]
    \centering
    \includegraphics[width=0.9\textwidth]{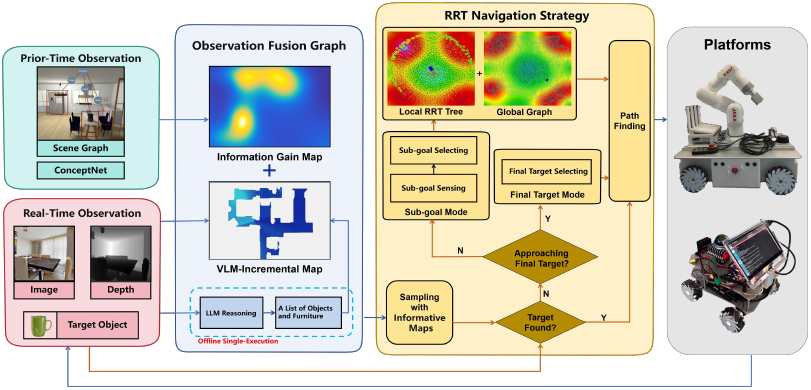}
    \caption{Overview of the proposed active search pipeline. The framework combines an IGM derived from the scene graph and commonsense knowledge with an incrementally updated VLM map informed by RGB-D observations and offline LLM reasoning. These two cues jointly guide the selection of local sub-goals, global targets, and the execution of the path.}
    \label{fig: pipline}
\end{figure*}
Another class of approaches \cite{yin2024sg,yokoyama2024vlfm} relies more heavily on local observation, combined with VLM-based semantic reasoning, to guide exploration and verification. These methods are more responsive to the current scene, but they can still exhibit unstable planning behavior in the presence of clutter, occlusion, or ambiguous context. The limitation becomes more severe for common small objects whose locations are not fixed and whose semantic context is highly variable, since the target may appear in different rooms and cannot always be reliably localized from local semantic cues alone. In such cases, observation-driven semantic guidance can undermine exploration efficiency by inducing revisits and inflating verification costs.

To address these limitations, we propose an active search framework that unifies historical knowledge and online semantic evidence within a single planning loop. Specifically, historical environmental experience is transformed into an IGM, which provides probabilistic global guidance toward regions that are more likely to contain the target under object context relations and commonsense priors. At the same time, current observations are processed by a VLM and fused over time into a VLM-SM, so that target-related evidence in the present scene can be incrementally accumulated rather than inferred from isolated frames. Building on these two semantic layers, we develop an IGV-RRT planner that couples prior target-discovery potential with online semantic support during tree expansion and sub-goal selection, enabling the robot to preserve efficient global search behavior while correcting outdated priors during execution. In this way, the proposed method explicitly addresses the mismatch between prior knowledge and the current environment, and provides a unified solution for active target search in temporally changing indoor scenes.

The contributions of this paper are as follows.
\begin{itemize} 
   \item We establish a dual-layer semantic mapping architecture comprising an IGM and a VLM-SM to jointly encode prior uncertainty and real-time semantic evidence in temporally changing environments.
   \item We propose an IGV-RRT navigation algorithm jointly guided by information gain and VLM scores for active target search.
   \item We implement the proposed framework on a real robotic platform and validate its effectiveness in real-world indoor environments.
\end{itemize}
The remainder of this paper is organized as follows. Section. \ref{sec: methodology} presents the proposed IGV-RRT method and its real-time planning mechanism driven by information gain and VLM scores. Section. \ref{sec: experiments} reports the experimental setup and evaluation results in both simulation and real-world environments. Finally, Section \ref{sec: conclusion} concludes the paper and discusses future work.

\section{Methodology}
\label{sec: methodology}
This study proposes an active object search framework that integrates prior scene knowledge and real-time semantic perception in a unified closed loop, as illustrated in Fig. \ref{fig: pipline}. The framework takes prior time observation and real-time observation as two inputs. From prior observations, a 3D scene graph and ConceptNet commonsense knowledge are used to construct an Information Gain Map for global guidance. From real-time RGB-D observations, target-related semantic measurements are inferred by the vision language model and incrementally fused into a VLM score map for online scene validation. Based on these two maps, the planner performs sampling, sub-goal selection, and path generation within the IGV-RRT framework. When local guidance is insufficient, the global graph further provides region-level guidance. Through repeated map updating, planning, execution, and verification, the system enables active target search in temporally changing indoor environments.
\subsection{Information Gain Map}
This paper represents the potential existence location of the target object $o_t$ as a probability density field over a 2D space $\mathcal{X}$, referred to as the IGM $P(x \mid o_t)$, where $x \in \mathcal{X}$. To construct this field, we first build a 3DSG $\mathcal{G}=(\mathcal{V},\mathcal{E})$ from multi-frame semantic observations in a unified world coordinate system \cite{hughes2022hydra}. The graph represents furniture and recognizable objects in the environment as semantic anchors. Each node $v_k\in\mathcal{V}$ corresponds to an anchor object instance $l_k$, and stores its semantic category, observation confidence $C_{conf}^{(k)}$, and geometric center location; because the IGM is defined over the ground plane, we project the instance center onto the ground plane and use this projection as the 2D anchor location, denoted as $\mu_k\in\mathbb{R}^2$. Here, the subscript $k\in\{1, \ldots, K\}$ indexes anchor object instances, and $K$ denotes the total number of anchors. The edge set $\mathcal{E}$ encodes topological or spatial context relations among anchor instances, providing structured contextual cues for subsequent commonsense scoring.

Given the constructed scene graph, we quantify commonsense relevance between the target category and each anchor instance using ConceptNet Numberbatch \cite{speer2017conceptnet} embeddings. Let the target category embedding be $\mathbf{v}_{target}\in\mathbb{R}^d$, where $d$ denotes the embedding dimensionality determined by ConceptNet Numberbatch; the category embedding of anchor instance $l_k$ be $\mathbf{v}_{cat}^{(k)}\in\mathbb{R}^d$, and its spatial-context embedding be $\mathbf{v}_{space}^{(k)}\in\mathbb{R}^d$. The cosine similarity is defined as
\begin{equation}
sim(u, v) = \frac{u \cdot v}{\|u\| \|v\|}, 
\end{equation}
to measure semantic proximity in the embedding space. The semantic association score $S(o_{t}, l_k)$ is then defined as a weighted combination of observation confidence and two similarity terms:
\begin{equation}
\begin{split}
S(o_{t}, l_k)
=\, C_{conf}^{(k)} \cdot \exp\!\Big(
&\, sim(\mathbf{v}_{target}, \mathbf{v}_{cat}^{(k)})+ \\
&sim(\mathbf{v}_{target}, \mathbf{v}_{space}^{(k)})
\Big).
\end{split}
\label{eq:semantic_score}
\end{equation}
where $C_{conf}^{(k)}$ serves as an observation-confidence factor that modulates the overall semantic association strength. Here, $sim(\mathbf{v}_{target}, \mathbf{v}_{cat}^{(k)})$ denotes the cosine similarity between the target category embedding and the anchor’s category embedding, and $sim(\mathbf{v}_{target}, \mathbf{v}_{space}^{(k)})$ denotes the cosine similarity between the target category embedding and the anchor’s spatial-context embedding. Based on the resulting discrete anchor set $\{l_k\}_{k=1}^{K}$ and their association scores, following \cite{wang2019semantic}, we employ a Gaussian Mixture Model (GMM) to extend these discrete anchors into a continuous probability density field over the entire space, yielding the IGM:
\begin{equation}
\label{eq:igm_gmm}
P(x\mid o_{t}) = \sum_{k=1}^{K} \phi_k \cdot \mathcal{N}(x \mid \mu_k, \Sigma_k),
\end{equation}
where the mixture weights are obtained by normalizing the association scores:
\begin{equation}
\phi_k = \frac{S(o_{t}, l_k)}{\sum_{j=1}^{K} S(o_{t}, l_j)},
\end{equation}
and each 2D Gaussian component is given by
\begin{equation}
\scalebox{0.85}{$\displaystyle
\mathcal{N}(x \mid \mu_k, \Sigma_k) = \frac{1}{2\pi \sqrt{|\Sigma_k|}} \exp\left( -\frac{1}{2} (x - \mu_k)^T \Sigma_k^{-1} (x - \mu_k) \right),
$}
\end{equation}
where $\mu_k\in\mathbb{R}^2$ is the projected geometric center of anchor instance $l_k$, and $\Sigma_k\in\mathbb{R}^{2\times 2}$ characterizes the spatial dispersion of the target around this anchor, and is specified in relation to the physical scale of the anchor object and the plausible spatial extent of the target distribution.

Note that the IGM, denoted as $P(x\mid o_{t})$, is computed from a snapshot of the environment at time $T$ and is subsequently used for planning at later times $t>T$. In real deployments, indoor scenes are time-varying; therefore, we treat the IGM as an informative but potentially biased prior, and do not assume it remains perfectly accurate over time.

\begin{figure*}[t]
    \centering
    \includegraphics[width=0.9\textwidth]{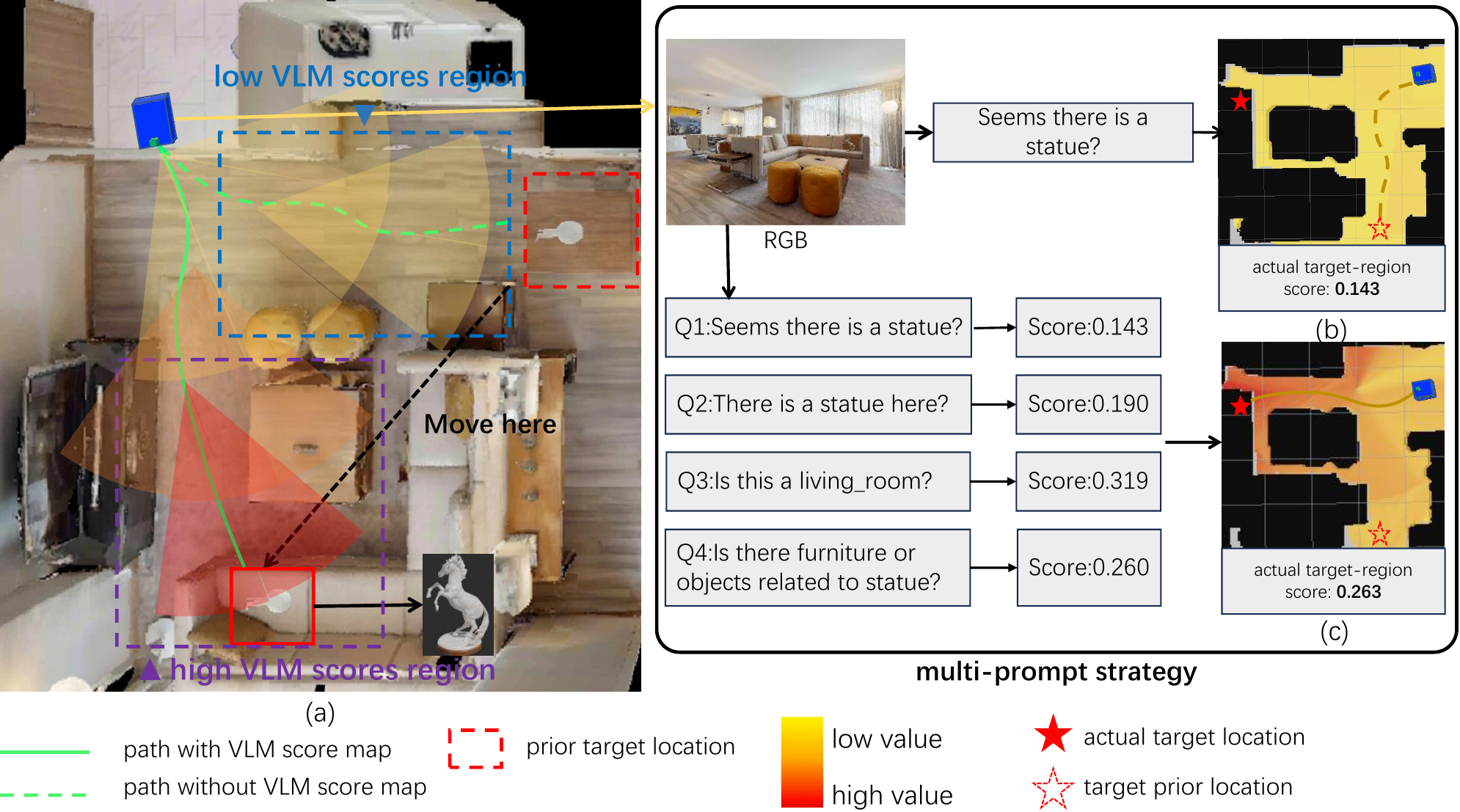}
    \caption{VLM correction and multi-prompting. (a) illustrates the corrective role of the VLM score map under a biased prior. When the prior indicates an incorrect target region, the robot is steered toward areas with higher VLM scores that reflect a higher likelihood of target presence, leading to effective progress toward the true object location.(b) and (c) highlight the impact of the prompting strategy. With a single prompt, the score response shows weak regional contrast and provides insufficient guidance for navigation. With multi-prompt querying, the score map exhibits stronger spatial discriminability, providing clearer guidance and enabling the robot to reach the target location more effectively. }
    \label{fig: VLM}
\end{figure*}

\subsection{VLM Score Map}
We construct a VLM Score Map using target-relevance scores inferred by a vision-language model~\cite{yokoyama2024vlfm}, and maintain a grid-based representation of the target-related score over spatial locations. The map is updated incrementally in a unified coordinate frame. For each grid cell $x$, its state is characterized by the accumulated confidence $C_t(x)$ and the semantic score $V_t(x)$, thereby fusing measurements collected at different times and viewpoints into a queryable evidence layer. The map is exposed through a persistent grid interface, and its outputs serve as one of the semantic inputs to downstream decision making.

Semantic observations are produced by BLIP-2 via image-text relevance evaluation. To improve the stability of BLIP-2 inference in object search, we employ an LLM in an offline stage to infer target-related contextual knowledge and convert it into a set of semantic queries. The generated knowledge captures contextual regularities associated with the target and is embedded into multi-granularity prompt templates, so that BLIP-2 can assess the current image not only with respect to direct target presence, but also through semantically related context. Based on this process, we construct a multi-prompt query set $\mathcal{T}=\{Q_1, Q_2, Q_3, Q_4\}$, where each $Q_k$ corresponds to a semantic query template at a different granularity and provides complementary information spanning direct existence descriptions and scene-context cues. For each image frame $I_t$, BLIP-2 outputs similarity scores between the image and each prompt, and we obtain the per-frame semantic observation score $v_{obs}$ by weighted aggregation,
\begin{equation}
v_{obs} = \sum_{k=1}^{4} w_k \cdot sim_{BLIP}(I_t, Q_k).
\end{equation}
Here, $w_k$ denotes the weight of each prompt, reflecting the contribution of different semantic cues to target relevance.

The effect of this multi-prompt design is illustrated in Fig.~\ref{fig: VLM}. As shown in Fig.~\ref{fig: VLM}(a), querying BLIP-2 with a single prompt often yields low spatial discriminability, which can be insufficient to reliably guide navigation in some cases. In contrast, the multi-prompt formulation in Fig.~\ref{fig: VLM}(b) produces a more distinctive response that better separates high-likelihood and low-likelihood regions for the target, thereby providing a stronger semantic signal to steer the robot toward the correct location.

To project semantic observations from the 2D image space to a 2D grid map, we adopt an instantaneous confidence model based on field-of-view geometry and perform recursive fusion at the grid level. Given the robot pose $\xi_t$ and a grid cell $x$, the instantaneous confidence of a single-frame observation for that cell is defined as
\begin{equation}
\scalebox{0.9}{$\displaystyle
c_{inst}(x, \xi_t) = 
\begin{cases} 
\left( \cos \left( \frac{\theta_{rel}}{\theta_{fov}/2} \cdot \frac{\pi}{2} \right) \right)^2 & \text{if } |\theta_{rel}| \le \frac{\theta_{fov}}{2} \\
0 & \text{otherwise}
\end{cases},
$}
\end{equation}
where $\theta_{rel}$ is the angular deviation of the cell center relative to the camera optical axis, and $\theta_{fov}$ is the horizontal field-of-view angle. The map state is represented by $(C_t(x),V_t(x))$ and updated incrementally through the following recursions,
\begin{equation}
C_t(x) = \frac{C_{t-1}^2(x) + c_{inst}^2(x, \xi_t)}{C_{t-1}(x) + c_{inst}(x, \xi_t)},
\end{equation}
\begin{equation}
V_t(x) = \frac{C_{t-1}(x) \cdot V_{t-1}(x) + c_{inst}(x, \xi_t) \cdot v_{obs}}{C_{t-1}(x) + c_{inst}(x, \xi_t)}.
\end{equation}
Here, $C_t(x)$ denotes the accumulated observation confidence of grid cell $x$ up to time $t$. $V_t(x)$ denotes the target relevance estimate of cell $x$, obtained by recursively fusing historical semantic estimates with the current observation. $v_{obs}$ is the target relevance score inferred from the current image frame and serves as the instantaneous semantic measurement used in the map update.

As a result, the VLM-SM incrementally fuses multi-temporal semantic observations on a unified grid, providing a queryable representation of semantic evidence along with its associated confidence.

\subsection{IGV-RRT Planning with Semantic and Information Gain}
The planning module employs IGV-RRT to perform online planning in continuous free space, and incrementally expands and maintains a sampling tree $\mathcal{T}$ at each planning cycle \cite{wang2024history}. The root of the tree is synchronized with the robot’s current state. During local expansion, feasible branches are generated under kinematic and collision constraints. The tree is further maintained through rewiring and root-rewiring mechanisms to preserve connectivity to the current root and consistency of path costs \cite{noreen2016optimal}, thereby enabling real-time responsiveness under continuous motion. In addition, the framework maintains a global graph composed of historical vertices and connectivity edges, which preserves a large-scale traversability structure and provides region-level guidance beyond the current local tree. Beyond this standard IGV-RRT framework, our key design is to incorporate the IGM and the VLM-SM as two complementary information sources within the same decision loop, such that both the expansion direction and sub-goal selection are jointly constrained by prior target-discovery potential and online semantic evidence.

\begin{figure}[t]
    \centering
    \includegraphics[width=1\linewidth]{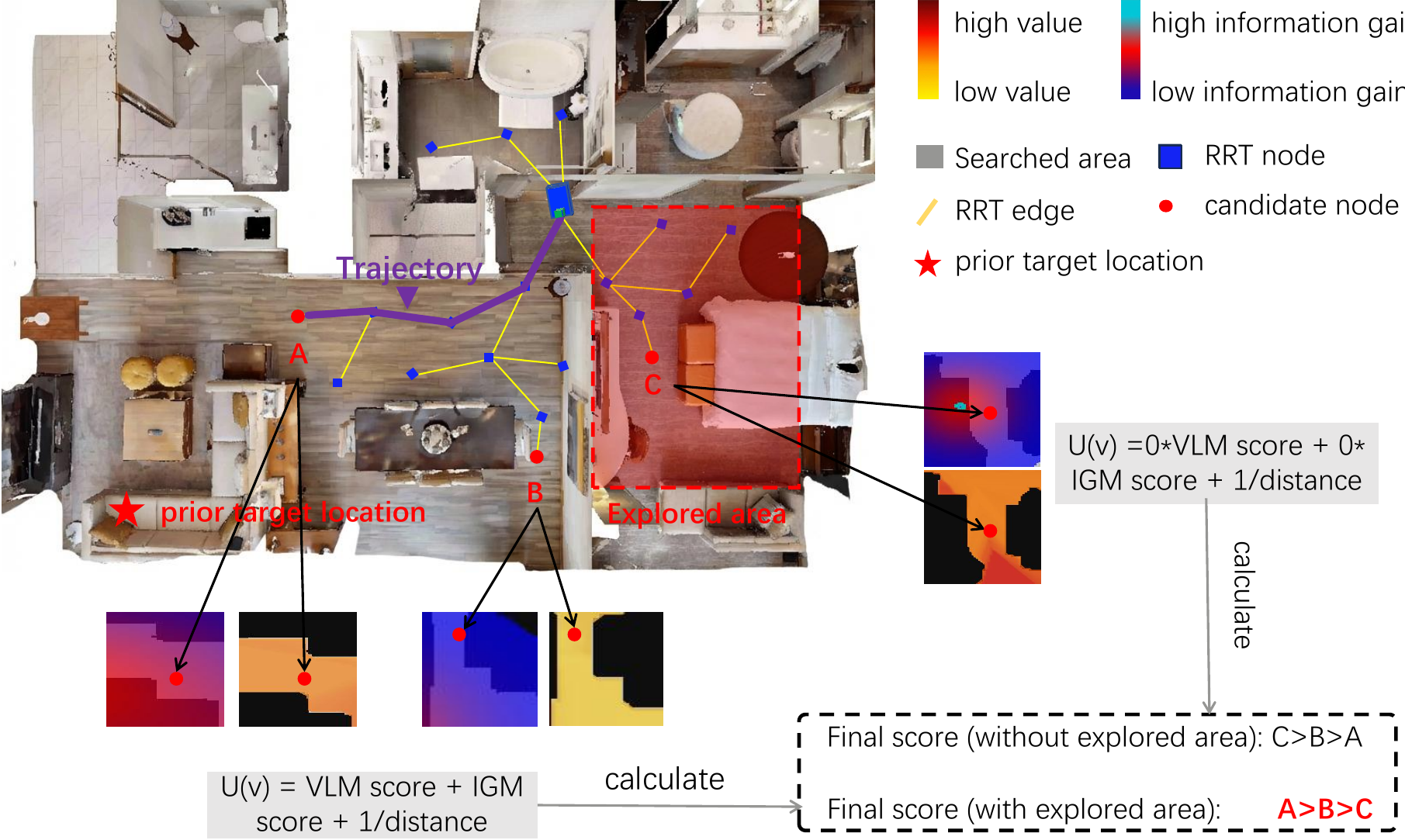}
    \caption{Utility-based frontier scoring with explored-region gating in IGV-RRT. The figure shows how IGV-RRT scores candidate frontiers by combining distance, IGM entropy, and VLM-SM evidence into a joint utility. An explored-region mask reduces the utility of previously observed areas to only the distance heuristic, encouraging selection of informative, unexplored frontiers. Without this gating, the planner may repeatedly choose already observed areas and miss the target. With the mask, it prefers the truly target approaching frontier A}
    \label{fig: explore}
\end{figure}

Specifically, the planner evaluates each candidate node $v$ using a joint utility that balances directional guidance, prior information gain, and online semantic evidence:
\begin{equation}
\begin{aligned}
U_{final}(v) =\;& \lambda_d \cdot (1 - D(v)) + \\
&\mathbb{I}(v \notin \mathcal{M}_{exp}) \cdot \left[\lambda_e \cdot E(v) + \lambda_s \cdot S(v)\right],
\end{aligned}
\end{equation}
where $D(v)$ denotes the normalized spatial distance from $v$ to the prior high-probability target point inferred from the IGM, and $\lambda_d \cdot (1-D(v))$ provides a weak but stable directional bias toward the prior target region. $\mathcal{M}_{exp}$ denotes the explored-region mask accumulated from the sensor field of view projected into the map frame, and $\mathbb{I}(\cdot)$ is the indicator function. When $v \in \mathcal{M}_{exp}$, both the prior gain term and the semantic support term are suppressed, so that the utility is determined only by the distance heuristic. The effect of explored-region gating on the utility evaluation is illustrated in Fig. \ref{fig: explore}.

To characterize the prior target-discovery potential around $v$, we compute the information-gain term over a local neighborhood $\Omega(v)$ as
\begin{equation}
E(v)=\sum_{x\in \Omega(v)} \mathbb{I}(x \notin \mathcal{M}_{exp}) \cdot \Big(-P(x\mid o_t)\log_2 P(x\mid o_t)\Big),
\end{equation}
which accumulates the entropy contribution of unexplored cells under the IGM prior. In the same neighborhood, the semantic support term is defined as
\begin{equation}
S(v)=\sum_{x\in \Omega(v)} \mathbb{I}(x \notin \mathcal{M}_{exp}) \cdot V_t(x),
\end{equation}
where $V_t(x)$ is the fused target-relevance estimate provided by the VLM score map. In our implementation, $\Omega(v)$ is taken as a circular neighborhood centered at $v$, so that $S(v)$ measures the accumulated online semantic evidence in the local region rather than the score of a single cell.

The weighting coefficients $\lambda_e$, $\lambda_s$, and $\lambda_d$ govern the relative contributions of prior information gain, online semantic evidence, and directional bias. Since $E(v)$ is derived from historical knowledge and may become biased under temporal scene changes, while $S(v)$ reflects current observations, $\lambda_s$ is assigned a larger value than $\lambda_e$. The coefficient $\lambda_d$ is set smaller than both so that the distance term remains auxiliary. At each planning cycle, the planner selects $v^*=\arg\max_v U_{final}(v)$ as the current sub-goal and generates a locally feasible trajectory toward $v^*$ for execution. As the robot moves and new observations arrive, both the tree structure and the utility values are updated continuously, forming a closed-loop process of expansion, evaluation, execution, and replanning. During execution, the system performs target detection and recognition using GroundingDINO \cite{liu2024grounding} and MobileSAM \cite{zhang2023faster}.

In cluttered indoor environments, IGV-RRT can still become locally trapped despite feasible paths. We add a stuck detection and escape mechanism: if the displacement within a preset time window falls below a threshold, the robot is considered stuck. The surrounding space is discretized into eight directions with an angular resolution of $45^\circ$, occupied cells in each directional neighborhood are counted, and the escape direction is selected by
\begin{equation}
\rho^* = \arg\min_{\rho_i \in \mathcal{R}} N_{\mathrm{obs}}(\rho_i)
\end{equation}
where $\mathcal{R}$ is the candidate direction set, $\rho_i$ is a candidate direction, and $N_{\mathrm{obs}}(\rho_i)$ is the number of occupied cells in its neighborhood. The robot executes a short motion along $\rho^*$ and resumes normal IGV-RRT planning once motion recovers.

In addition, when the number of valid local sub-goals in the current region becomes insufficient, the planner invokes the global graph to provide region-level guidance and steer the robot out of the current area. In this sense, the escape mechanism handles local deadlock at a specific blocked position, whereas the global graph addresses regional stagnation caused by insufficient local guidance. This design improves robustness in cluttered indoor scenes without introducing significant planning overhead.

Overall, the system performs target object search under temporal scene changes by explicitly addressing the mismatch between the prior construction time $T$ and the execution observation time $t>T$. The IGM provides a coarse global bias that allocates the search budget to regions that are more consistent with commonsense under the anchor-object context, thereby preventing the process from degenerating into purely geometric coverage. The VLM-SM incrementally accumulates online semantic observations via confidence-weighted fusion, enabling prior validation at execution time and reducing the misleading effect of obsolete priors. Building on this, IGV-RRT incorporates the above prior and evidence into utility evaluation to guide tree expansion and sub-goal selection, while an explored-region mask suppresses revisits, thus enabling kinematically feasible, target-directed exploration.

\section{Experiments and Evaluation}
\label{sec: experiments}
\subsection{Simulation Experiments}
We evaluate our method on the HM3D \cite{ramakrishnan2021habitat} simulator under a temporal-change ObjectNav setting, where object locations vary over time. Based on HM3D, we augment each scene by importing additional objects, such as statues and vases, and construct a new task set in which these imported objects serve as navigation targets. We decouple prior construction from policy execution. Specifically, the IGM is built and frozen from a scene snapshot at time $T$, while during execution, the imported objects may be moved to simulate temporal rearrangements, thereby inducing a mismatch between the prior and the current environment. This benchmark is intentionally designed because when the prior remains accurate, a static IGM alone can already provide sufficient guidance, making the marginal benefits of online semantic correction and revisit suppression difficult to isolate. By contrast, the temporal mismatch setting offers a controlled testbed in which the corrective effect of online semantic evidence and the efficiency gain brought by revisit suppression can be evaluated more clearly. Under this setting, we compare our method with several representative state-of-the-art approaches.

\begin{table}[!htbp]
  \centering
  \caption{Simulation results on HM3D under temporal changes}
  \renewcommand{\arraystretch}{1.15}
  \setlength{\tabcolsep}{8pt}
  \begin{tabular}{lcc}
    \toprule
    \multirow{2}{*}{Method} & \multicolumn{2}{c}{\textbf{HM3D}} \\
    \cmidrule(lr){2-3}
    & \textbf{SR $\uparrow$} & \textbf{SPL $\uparrow$} \\
    \midrule
    CoW \cite{gadre2023cows} & 15.8  & 7.36 \\
    ZSON \cite{majumdar2022zson} & 26.54 & 9.03 \\
    PSL \cite{sun2024prioritized} & 40.18 & 17.68 \\
    VLFM \cite{yokoyama2024vlfm} & 49.76 & 27.43 \\
    OneMap \cite{busch2025one} & 54.38 & 33.23 \\
    \rowcolor{gray!12}
    \textbf{IGV-RRT (Ours)} & \textbf{64.91} & \textbf{39.28} \\
    \bottomrule
  \end{tabular}
  \label{table: DB}
\end{table}

As shown in Table \ref{table: DB}, IGV-RRT achieves the best performance among all compared methods. We report two standard navigation metrics, namely Success Rate (SR) and Success weighted by inverse Path Length (SPL). SR is computed as the fraction of episodes in which the robot successfully reaches the target. SPL is computed as the average of success weighted by the ratio between the shortest-path distance to the goal and the actual path length executed by the robot, thereby jointly reflecting task completion and navigation efficiency. Under these two metrics, IGV-RRT attains the highest SR and SPL, indicating that the proposed method not only improves the probability of finding the target object but also enables more efficient navigation once a feasible search direction is established. 

Beyond the quantitative metrics, we visualize the navigation process on different tasks.
\begin{figure}[t]
    \centering
    \includegraphics[width=1\linewidth]{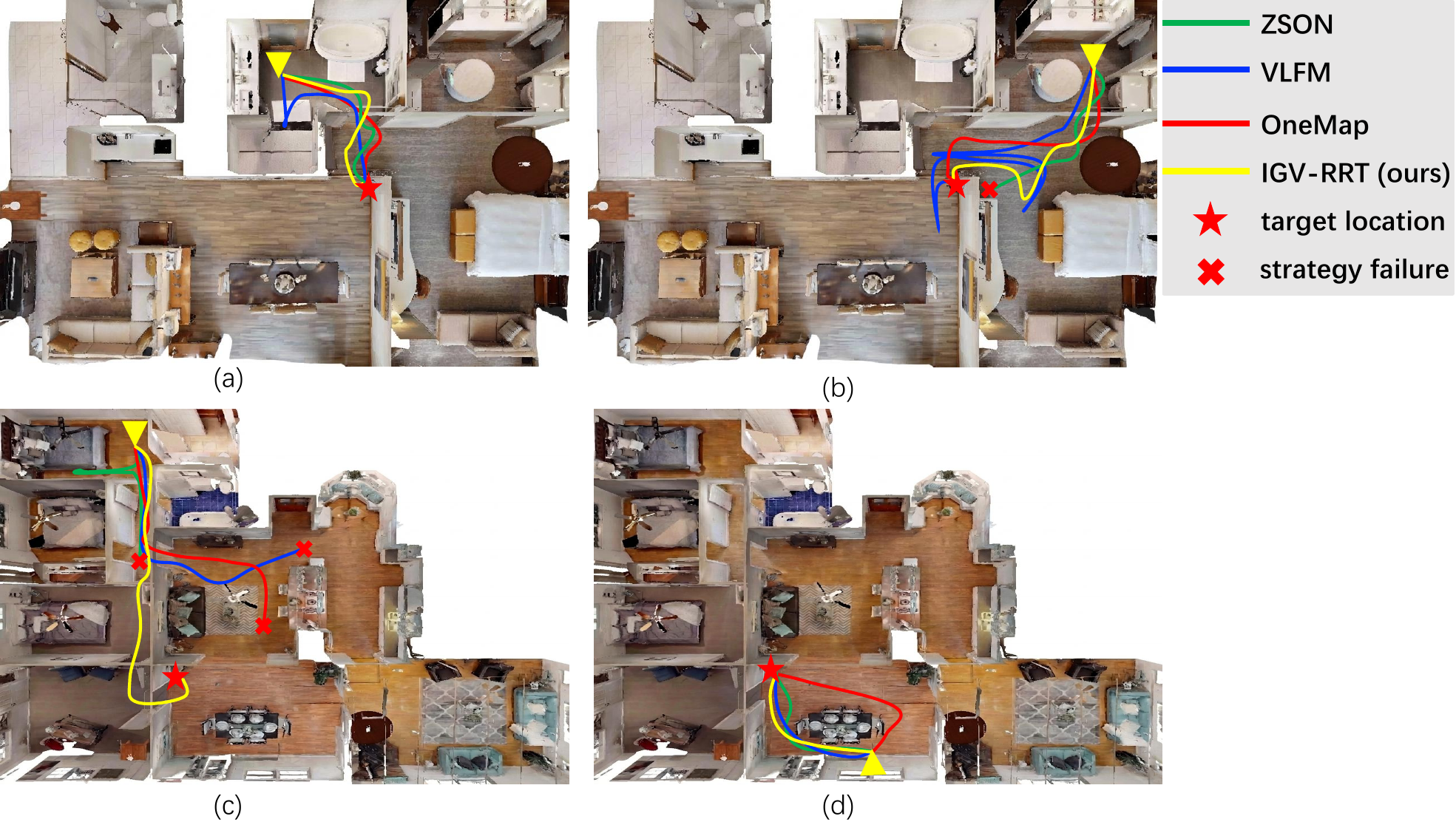}
    \caption{Comparison of navigation strategies across different tasks. The figure presents the navigation outcomes of different strategies over different tasks in different scenes. Red crosses denote failure locations where navigation exceeded the maximum step limit.}
    \label{fig: experiment}
\end{figure}
 Fig. \ref{fig: experiment} compares the navigation performance of our method with that of the compared methods across different tasks in different scenes. As shown in  Fig. \ref{fig: experiment}(a) and  Fig. \ref{fig: experiment}(d), when the initial position is relatively close to the target object, most methods can successfully reach the target by relying on semantic cues. However, when the robot starts farther from the target, the limitation of purely semantic guidance becomes more evident. For objects such as a vase, whose semantic association with specific rooms or surrounding furniture is relatively weak, the target may plausibly appear in many different locations within a home. As a result, the compared methods are more prone to navigation failure or repetitive trajectories, as illustrated in  Fig. \ref{fig: experiment}(b) and  Fig. \ref{fig: experiment}(c). In contrast, our method leverages the IGM to provide global guidance, progressively steering the robot toward the region where the target is more likely to be located. After entering that region, the robot utilizes real-time visual observations to correct its prior bias and ultimately navigate to the true target location.

\subsection{Ablation Study}
To evaluate the roles of individual components specifically under prior mismatch, we conduct an ablation study on the same benchmark built on the modified HM3D scenes. This setting provides a controlled testbed for quantifying the corrective effect and efficiency contribution of each component. Specifically, we compare four strategies that all use the Information Gain Map as the base signal, but differ in whether the VLM-based semantic map is incorporated and whether explored-region suppression is enabled. The results are summarized in Table \ref{table: XR}.
\begin{figure*}[!htbp]
    \centering
    \includegraphics[width=0.9\textwidth]{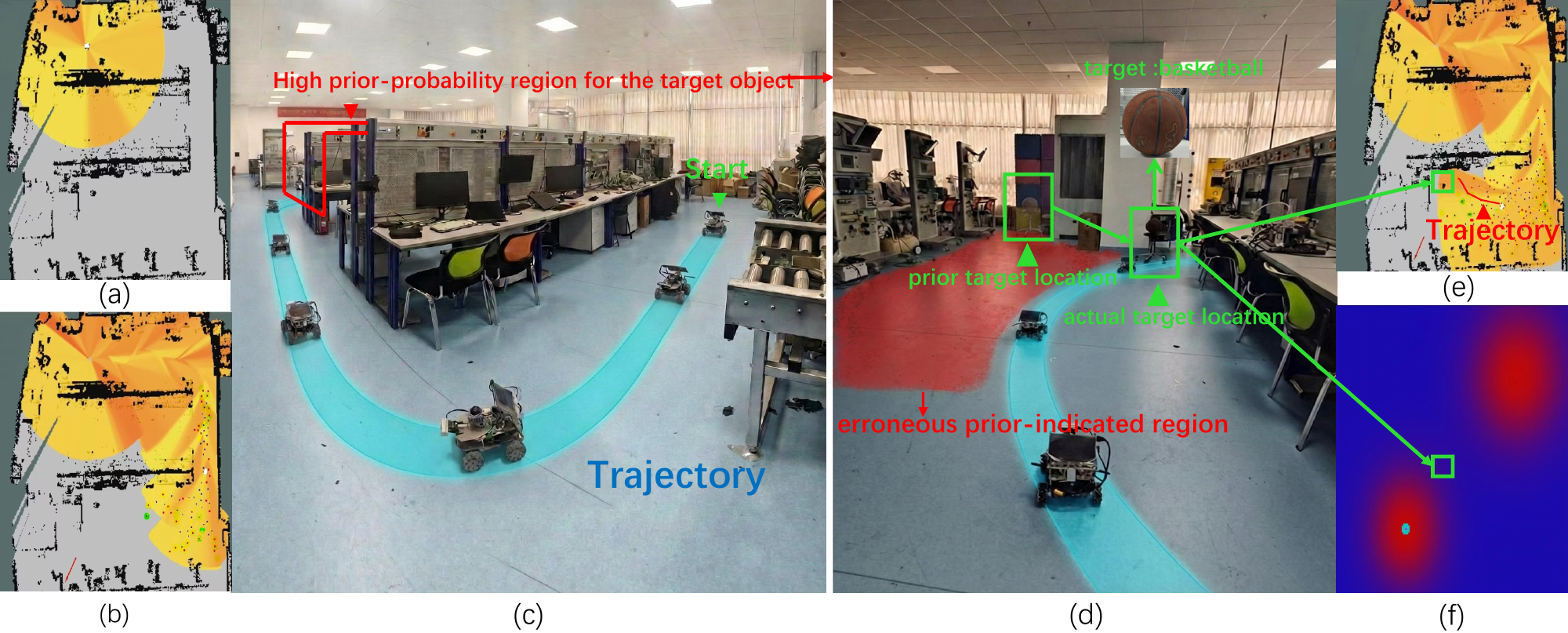}
    \caption{Real-world navigation using IGV-RRT. (a)-(c) show that, in the early stage, the robot is rapidly driven by the prior IGM toward high probability regions where the target is likely to exist. During this stage, even when the VLM score map varies across regions, it does not dominate the motion unless the score differences become sufficiently pronounced. (d) and (e) indicate that, after entering the high probability region, the robot increasingly follows the VLM score map and progresses toward the true target location. (f) further reveals the mismatch and bias that can arise from the prior IGM.}
    \label{fig: realword}
\end{figure*}

\begin{table}[!htbp]
  \centering
  \caption{Ablation results on the temporal-change benchmark}
  \renewcommand{\arraystretch}{1.15}
  \setlength{\tabcolsep}{10pt}
  \begin{tabular}{lcc}
    \toprule
    \textbf{Strategy} & \textbf{SR $\uparrow$} & \textbf{SPL $\uparrow$} \\
    \midrule
    (a) IGM only & 50.87 & 28.03 \\
    (b) IGM + explored-region & 53.92 & 33.89 \\
    (c) IGM + VLM-SM & 56.28 & 34.95 \\
    \rowcolor{black!6}
    (d) \textbf{Ours} & \textbf{64.91} & \textbf{39.28} \\
    \bottomrule
  \end{tabular}
  \label{table: XR}
\end{table}

Table \ref{table: XR} suggests a complementary interplay among the components. Using the information gain map alone under temporal mismatch is susceptible to outdated prior bias, which can steer sub-goal selection toward prior-favored but incorrect regions and incur unnecessary verification. In this setting, even without introducing VLM-based online semantic evidence, enabling explored-region suppression alone can still yield a modest improvement in success rate and path efficiency. This is because explored-region suppression prevents the robot from repeatedly revisiting previously explored areas, thereby allowing it to allocate more search effort to unexplored regions. As a result, the likelihood of discovering the target object is increased, while redundant path traversal is reduced. Adding the VLM-based semantic map introduces online semantic evidence that corrects prior induced bias during execution, enabling faster recovery from misleading priors and improving both reliability and efficiency. Enabling explored region suppression is most effective when paired with sufficiently informative guidance, where it reduces revisits and loop-like traversal, promotes forward progress, and improves path efficiency. Combining both VLM-based correction and revisit suppression in the full strategy, therefore, provides the most robust overall performance under temporal object displacement, typically achieving the best or near-best success rate and success weighted by path length jointly.

\begin{figure}[!htbp]
    \centering
    \includegraphics[width=1\linewidth]{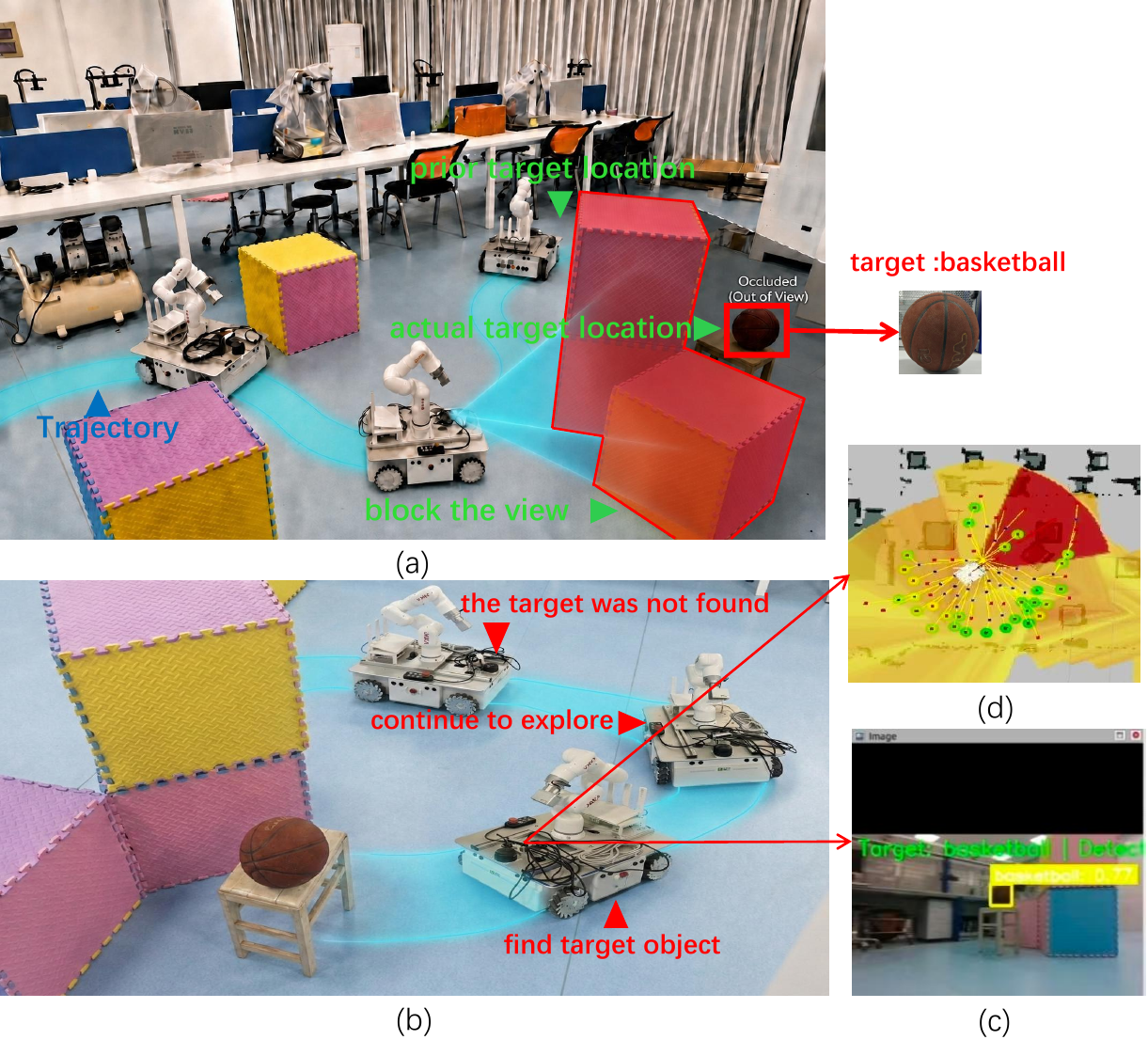}
    \caption{Prior-biased search under occlusion. (a) and (b) show the robot reaching a prior-biased goal under obstacle occlusion, failing to detect the target, then continuing exploration to recover the true object. (c) and (d) depict the re-detection phase via the GroundingDINO view and the VLM-SM.}
    \label{fig: experiment2}
\end{figure}

\subsection{Real-World Experiments}

Real-world experiments are conducted on two ROS mobile platforms: a Wheeltec R550 and a VSAI-SFM AGV. The R550 is equipped with an Orin Nano 4G as the onboard computing unit and an Intel RealSense D435i RGB-D camera for visual perception and depth sensing. The VSAI-SFM AGV is equipped with an ARMv8 CPU, a D435i depth camera, and an RPLIDAR A3M12 LiDAR for perception and sensing. The compute side is a desktop workstation equipped with an NVIDIA GeForce RTX 5060 Ti GPU and a 12th Gen Intel(R) Core(TM) i7-12700KF CPU, serving as the ROS master to run perception, semantic inference, and planning, while the robot side operates as a ROS slave for motion execution and state feedback. This master–slave deployment decouples high-load computation from onboard actuation, helping maintain both online planning throughput and control-loop stability.

Fig. \ref{fig: realword}(a) shows that the robot starts from the initial position and that the early trajectory moves toward the high-likelihood region indicated by the prior information gain, providing global guidance. Fig. \ref{fig: realword}(b) shows that after the robot enters this region, the trajectory does not proceed to the outdated, incorrect prior-indicated location; instead, it is progressively adjusted as online semantic observations are updated, where the VLM-SM provides fine-grained guidance. The trajectory then further departs from the prior-biased location, turns toward the true target, and finally converges near the actual object position. This figure directly illustrates the full execution process from the prior-guided approach to online correction, confirming the effectiveness and robustness of the proposed method in real, temporally changing environments.

Fig. \ref{fig: experiment2}(a) shows that while moving toward an incorrect prior goal, obstacle occlusion prevents real-time visual observations from correcting the bias accurately, so the target is not detected. Fig. \ref{fig: experiment2}(b) shows that after reaching this wrong prior goal, the robot confirms the absence, continues exploration, and detects the occluded target during subsequent exploration. This figure illustrates the method’s effectiveness and feasibility under prior mismatch and occlusion.

\section{Conclusion and Future Work}
\label{sec: conclusion}
We presented a probabilistic planning framework for active target search in indoor environments where target objects may be relocated over time, making purely static assumptions insufficient for reliable navigation. The framework integrates two complementary semantic cues: an IGM derived from scene-graph and commonsense reasoning to provide coarse global guidance, and an incrementally updated VLM-SM to inject real-time semantic evidence for local validation. Building on these maps, we proposed IGV-RRT, which unifies information gain, VLM scores, and navigation cost in a joint utility for tree expansion and sub-goal selection, while an explored-region mask suppresses revisits to improve search efficiency. Simulation and real-world experiments show that the proposed method achieves robust target-directed exploration and improves both success rate and path efficiency in challenging indoor scenes with object relocation.

plFuture work will focus on making the IGM updateable for long-term autonomy: the robot will detect persistent scene changes online and revise the IGM accordingly, keeping long-term guidance consistent with the evolving environment.

\addtolength{\textheight}{-12cm}   






\bibliographystyle{ieeetr}
\bibliography{ref}

\end{document}